\documentclass{article}

\usepackage[final]{bdl_2021_camera_ready}

\usepackage[utf8]{inputenc} %
\usepackage[T1]{fontenc}    %
\usepackage{hyperref}       %
\usepackage{url}            %
\usepackage{booktabs}       %
\usepackage{amsfonts}       %
\usepackage{nicefrac}       %
\usepackage{microtype}      %
\usepackage{xcolor}         %

\usepackage{url}

\hypersetup{
    colorlinks=True,
    linkcolor=darkblue,
    citecolor=darkblue,
    filecolor=darkblue,
    urlcolor=darkblue,
    }
\usepackage{amsfonts}       %
\usepackage{mathtools}      %
\usepackage{amsmath}
\usepackage{multirow}       %
\usepackage{subcaption}     %
\usepackage{placeins}       %
\usepackage{todonotes}
\usepackage{wrapfig}
\usepackage{amssymb}
\usepackage{adjustbox}      %
\usepackage[all]{nowidow}   %
\usepackage{xspace}         %
\usepackage{tikz}
\usetikzlibrary{bayesnet}
\usetikzlibrary{arrows}
\usepackage{amsthm}         %
\usepackage{xfrac}          %

\usepackage{bbm}
\usepackage{natbib}
\usepackage{color}
\usepackage{nkj,nkj_e}
\usepackage{algorithm}
\usepackage{algorithmicx}
\usepackage[noend]{algpseudocode}
\usepackage{bbold}

\usepackage[capitalize]{cleveref}
\crefname{appendix}{App.}{Apps.}        %
\crefname{section}{§}{§§}               %
\Crefname{section}{Section}{Sections}
\crefname{prop}{Proposition}{Propositions}
\crefname{definition}{Definition}{Definitions}
\crefname{hyp}{Hypothesis}{Hypotheses}
\creflabelformat{equation}{#2#1#3}      %

\usepackage{xcolor}
\definecolor{brown}{HTML}{8F3200}
\definecolor{green}{HTML}{005E37}
\definecolor{blue}{rgb}{0,0.08,0.45}
\definecolor{red}{HTML}{8B0000}
\definecolor{lik_proposed}{HTML}{125E8A}
\definecolor{purple}{HTML}{A14DA0}
\definecolor{navy}{HTML}{000080}
\definecolor{darkblue}{rgb}{0,0.08,0.45}

\usetikzlibrary{tikzmark}
\tikzset{mycircled/.style={circle,draw,inner sep=0.1em,line width=0.04em}}

\newcommand{\x}{\boldsymbol{x}}

\newcommand{\g}{\boldsymbol{g}}

\newcommand{\E}{\mathbb{E}}
\newcommand{\p}{p_{\theta}}
\newcommand{\q}{q_{\phi}}

\newif\ifshowComments
\showCommentstrue

\title{
Mixture-of-experts VAEs can disregard variation in surjective multimodal data}

\author{
\resizebox{\linewidth}{!}{
\begingroup
\tabcolsep = 2.0pt %
\begin{tabular}[t]{c}
  Jannik Wolff\thanks{Correspondence to: \texttt{wolff.jannik@icloud.com}}
  \thanks{Part of the work was done at SAP AI Research.}
  \\
  \normalfont{TU Berlin}
\and
  Tassilo Klein, Moin Nabi
  \\
  \normalfont{SAP AI Research}
\and
  Rahul G. Krishnan\thanks{Part of the work was done at Massachusetts Institute of Technology and Microsoft Research.}
  \\
  \normalfont{University of Toronto}
\and
  Shinichi Nakajima
  \\
  \normalfont{TU Berlin}
\end{tabular}
\endgroup
}
}

\begin{document}

\setkeys{Gin}{draft=False}

\linepenalty=1000

\maketitle

\begin{abstract}
Machine learning systems are often deployed in domains that entail data from multiple modalities, for example, phenotypic and genotypic characteristics describe patients in healthcare.
Previous works have developed multimodal variational autoencoders (VAEs) that generate several modalities.
We consider surjective data, where single datapoints from one modality (such as class labels) describe multiple datapoints from another modality (such as images).
We theoretically and empirically demonstrate that multimodal VAEs with a mixture of experts posterior can struggle to capture variability in such surjective data.
\end{abstract}

\section{Introduction}

\begin{wrapfigure}{r}{0.3\linewidth}
\vspace{-0.5cm}
\centering
\includegraphics[width=0.7\linewidth]{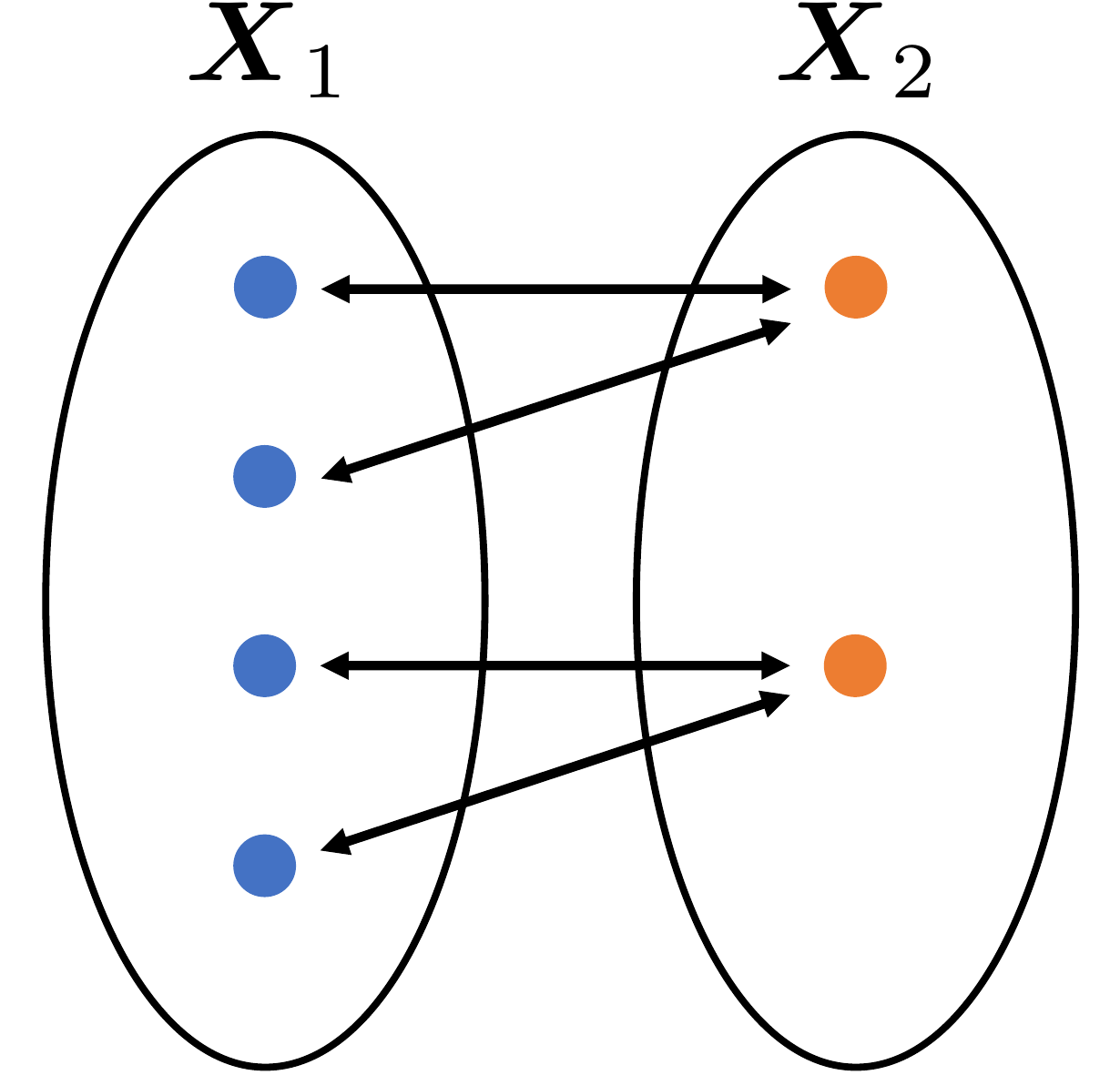}
\caption{
\textbf{\textit{Surjective} data.}
$X_1$ and $X_2$ depict exemplary modalities.
The mapping from the second to the first modality is surjective.}
\label{fig:surjective_data}
\vspace{-0.5cm}
\end{wrapfigure}

Many datasets entail a surjective mapping between modalities (\cref{fig:surjective_data}, \textit{``one-to-many data''}).
That is, an instance from one modality may correspond to several instances from another modality.
For example, many computer vision datasets contain labels, attributes, or text data that describe sets of images \citep{lecun1998mnist,nilsback2008automated,krizhevsky2009learning,deng2009imagenet,wah2011caltech,liu2015faceattributes,xiao2017fashion}.
Note that \textit{``one-to-one data''} such as image/caption pairs can become surjective when using data augmentation, e.g., random horizontal flipping of images.
Incorporating further modalities can also invoke surjectivity.

Multimodal VAEs maximize a bound on the joint density of several modalities and can thereby learn to generate any modality from any conditioning modality \citep{suzuki2016joint}.
For some multimodal VAEs, this bound contains a factor that represents the likelihood of one modality given another modality.
We will show that such a factor in the objective function can lead to solutions that disregard heterogeneity within a modality.
For example, we demonstrate that samples from models with a mixture of experts posterior such as the MMVAE \citep{shi2019variational} can have a bias towards the class mean of the observed datapoints for a given modality.

\section{Method}
\label{sec:method}

Let $\bfX = \{\{\bfx_{m}^{(n)}  \}_{m=1}^M\}_{n=1}^N$ be a training set with several modalities, where $m$ and $n$ represent the modality and the sample index, respectively.
We consider a multimodal VAE with a generative model
\begin{align}
\bfg & \sim p_{\theta}(\bfg),
\notag\\
\bfx_m & \sim p_{\theta}(\bfx_m | \bfg)  \qquad \mbox{for }m = 1, \ldots, M,
\label{eq:GenerativeGeneral}
\end{align}
and an inference model
\begin{align}
\bfg \sim q_{\phi}(\bfg | \{\bfx_m\}_{m=1}^M).
\label{eq:InferenceGeneral}
\end{align}

Assume that the generative model \eqref{eq:GenerativeGeneral} is a parametric model, e.g., Gaussian,
\begin{align}
p_{\theta}(\bfx_m | \bfg) = f_m(\bfx_m | \bftau_m(\bfg; \bftheta) ),
\label{eq:Generative}
\end{align}
with the parameters $\{\bftau_m\}$, e.g., means and covariances, defined as a function of $\bfg$ and (typically) neural networks weights $\bftheta$.
Assume that the inference model \eqref{eq:InferenceGeneral} is defined as a finite mixture with parameters $\bfkappa_m$ indicating mean and covariance for mixture component $r_m$ (as in the MMVAE \citep{shi2019variational}, for example):
\begin{align}
q_{\phi}(\bfg | \{\bfx_m\}_{m=1}^M)
& = \frac{1}{M} \sum_{m=1}^M   q_{\phi}(\bfg | \bfx_m)
= \frac{1}{M} \sum_{m=1}^M   r_m(\bfg |\bfkappa_m(\bfx_m; \bfphi) ).
\notag
\end{align}

Without loss of generality, we assume that $\bfx_M$ is the label modality, and let
$\bfS_c = \{n \mid \bfx_M^{(n)} = c \}$ be the set of indices of the samples belonging to the label $c \in \{ 1, \ldots, C\}$.
 We consider a maximization problem given the following objective function:
 \begin{align}
 L_m(\bftheta, \bfphi; \bfX) &\equiv \sum_{n=1}^N  \int   r_M(\bfg |\bfkappa_M(\bfx_M^{(n)}; \bfphi) ) \log f_m(\bfx_m^{(n)} | \bftau_m(\bfg; \bftheta) ) d\bfg,
 \label{eq:Objective}
 \end{align}
 which is an ELBO for
 \begin{align}
 \log p(\bfx_m | \bfx_M)
 &= 
 \log
 \int  q_{\phi}(\bfg | \bfx_M)  p_{\theta}(\bfx_m | \bfg)  d\bfg
\geq 
 \int  q_{\phi}(\bfg | \bfx_M)  \log p_{\theta}(\bfx_m | \bfg)  d\bfg
 = L_m(\bftheta, \bfphi; \bfX).
 \notag
 \end{align}
 Importantly, the MMVAE \citep{shi2019variational} relies on term \eqref{eq:Objective} for learning data translation ability from $\bfx_M$ to $\bfx_m$.
 Specifically, the authors used stratified sampling for training\footnote{Moving $\Sigma_m$ into the $\log$ in \cref{eq:moe_stratified_sampling} would imply a tighter bound. However, the model may then weigh the experts differently w.r.t. to their gradients, which can disproportionally favor the representation of single modalities at the expense of learning structure across all modalities.}, which implies that \cref{eq:Objective} and term \tikzmarknode[mycircled]{t1}{1} from \cref{eq:moe_stratified_sampling} are related:
\begin{align}
\begin{split}
 &\log \p(\{\bfx_m\}_{m=1}^{M}) \geq
\frac{1}{M}\sum_{m=1}^M \E_{\q(\g|\x_m)} \big[ \log 
    \frac{\p(\g, \{\bfx_m\}_{m=1}^{M})}{\q(\g|\{\bfx_m\}_{m=1}^{M})} \big]\\
= &\frac{1}{M}
\Bigg(
\sum_{m=1}^{M-1}
\left(
\E_{\q(\g|\x_{m})} \big[ \log \frac{\p(\g, \{\bfx_m\}_{m=1}^{M})}{\q(\g|\{\bfx_m\}_{m=1}^{M}))} \big]
\right)
+
\E_{\q(\g|\x_{M})} \big[ \log 
\frac{\p(\g)}{\q(\g|\{\bfx_m\}_{m=1}^{M})}
\big]\\
&+
\sum_{i=1}^M
\underbrace{\E_{\q(\g|\x_{M})} \big[ \log \p(\bfx_i|\g) \big]
}_{\tikzmarknode[mycircled]{t1}{1}}
\Bigg)
\label{eq:moe_stratified_sampling}
\end{split}
\end{align}

The following theorem holds:
\begin{theorem}
Assume a training set $X = \{\bfx_m^{(n)}\}_{n \in S_c}$ which belong to the same label, i.e., $\bfx_M^{(n)} = c, \forall n \in S_c$,
and there exists $\widehat{\bftheta}$ such that $\bftau_m(\bfg; \widehat{\bftheta})$ is a constant with respect to $\bfg$ and the maximum likelihood estimator of the parametric model $f_m(\bfx_m | \bftau_m(\bfg; \bftheta) ) $ for the training data.
Then, for any $\bftheta$, $\bfphi$, it holds that 
\begin{align}
 L_m (\widehat{\bftheta}, \bfphi; \bfX) \geq  L_m(\bftheta, \bfphi; \bfX).
 \label{eq:Optimum}
\end{align}
\label{theorem:main}
\end{theorem}
(Proof)
Since we assume that $\bfx_M^{(n)} = c$ for all $n \in \bfS_c$,
the inferred distribution  for $\bfg$ is the same for all $n$, i.e., $\widetilde{r}_M(\bfg) =  r_M(\bfg |\bfkappa_M(\bfx_M^{(n)}; \bfphi) )$.
For any such inference model $\widetilde{r}_M(\bfg) $, the objective is upper-bounded by
 \begin{align}
 L_m(\bftheta, \bfphi; \bfX) &= \int   \widetilde{r}_M(\bfg; \bfphi  )  \left( \sum_{n=1}^N  \log f_m(\bfx_m^{(n)} | \bftau_m(\bfg; \bftheta) )\right) d\bfg\\
 &\leq \int   \widetilde{r}_M(\bfg; \bfphi  )  \left( \sum_{n=1}^N  \log f_m(\bfx_m^{(n)} | \widehat{\bftau}_m )\right) d\bfg
\notag
 \end{align}
 with the maximum likelihood estimator $\widehat{\bftau}_m$ for the parametric model $ f_m$ given the training set $\{\bfx_m^{(n)}\}_{n=S_c}$.
The assumed existence of $\widehat{\bftheta}$ such that $\bftau_m(\bfg; \widehat{\bftheta}) = \widehat{\bftau}_m $ leads to Eq.~\eqref{eq:Optimum}.
\QED

Intuitively, consider a single class: $c \in \{1\}$.
Let $\p(\x_{m}|\g)$ be Gaussian with diagonal covariance, where $\g \sim \q(\g|\x_M)$.
\Cref{theorem:main} implies the existence of an upper bound where the mean parameter from $\p(\x_{m}|\g)$ always coincides with the mean from $\{\x_m^n\}_{n \in S_c}$ for any $\g$.
This solution is invariant to $\g$ because $\x_M$ does not carry information about across-datapoint variability in $\x_m$.
In other words, the solution maximizes the likelihood of the training data $\{\bfx_m^{(n)}\}_{n=S_c}$ with a single Gaussian distribution.
That is, the mean parameter minimizes the distance to all datapoints from modality $m$ simultaneously:
the model captures the mean of the target distribution -- not its variability.

\section{Experiments}
\label{sec:experiments}

We create a synthetic dataset (inspired by \citet{johnson2016composing}) with modality $\x_1 \in \mathbb{R}^2$ and label modality $\x_2 \in \{ 0,1\}$.
We implement the MVAE \citep{wu2018multimodal} and MMVAE \citep{shi2019variational}.
The latent distributions are isotropic Gaussian.
The generative distributions are isotropic Gaussian for the first modality and categorical for the second modality.

\begin{figure}[t]
\resizebox{\linewidth}{!}{
  \centering
  \includegraphics[width=\linewidth]{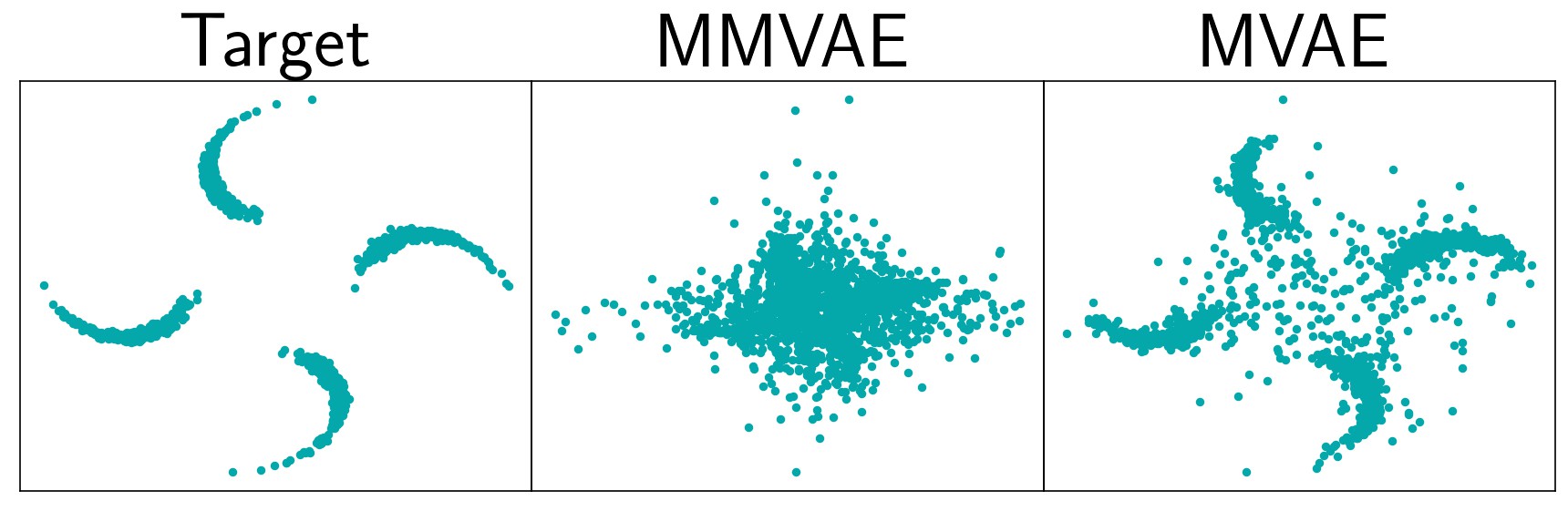}
  \includegraphics[width=\linewidth]{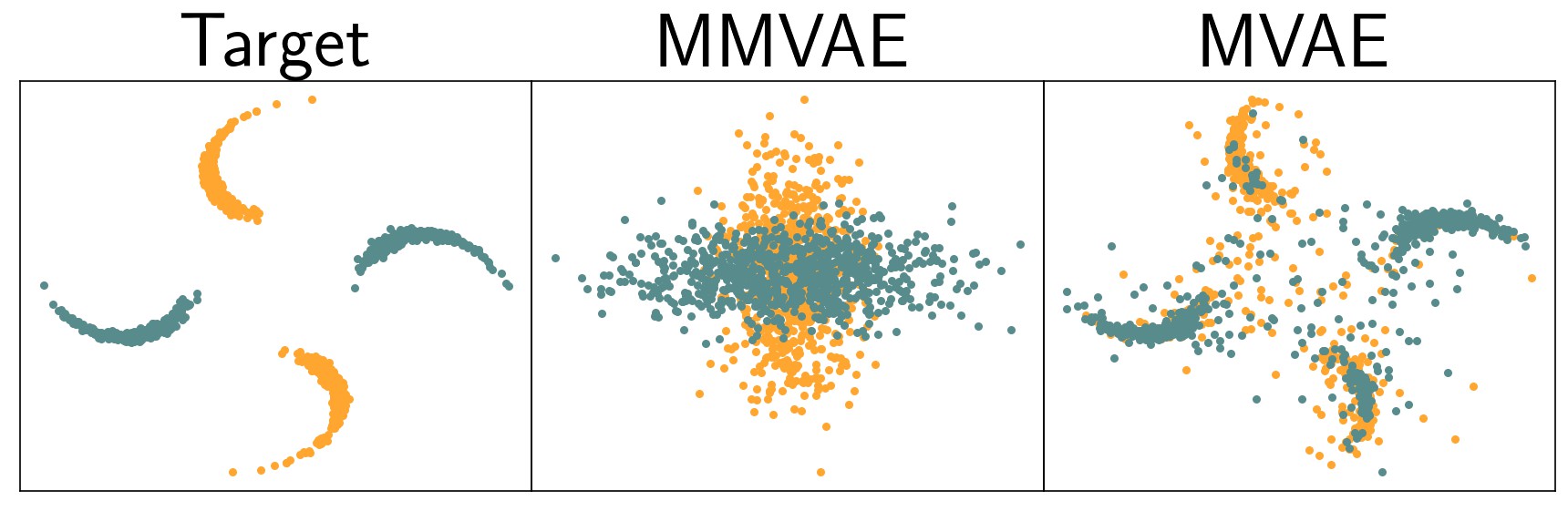}
}
\caption{\textbf{Generated samples for the first modality.}
Left: using samples from $p(\g)$.
Right: using samples from $q(\g|\x_2)$, where $\x_2$ are class labels (yellow or green).
}
\label{fig:model_samples}
\end{figure}

For the MMVAE, \cref{fig:model_samples} supports our argument that samples for the first modality tend towards the mean of the observed datapoints (for the same class).
The MVAE does not suffer from this problem, possibly because the MVAE's objective function does not contain the factor $p(\x_1|\x_2)$ (\cref{sec:app_mvae}).
\Cref{sec:app_exps} visualizes the latent spaces, which are two-dimensional to avoid possible obfuscation from dimensionality-reduction techniques.

\FloatBarrier
\section{Conclusion}

We show that multimodal VAEs with a mixture posterior can struggle to capture heterogeneity in surjective data.
This finding implies that practitioners should closely consider the type of data when training such models:
for example, data augmentation may not be beneficial since this procedure often promotes surjectivity.
Future work may investigate possible solutions, e.g., by considering models that do not maximize $p(\x_m|\x_{M\neq m})$ explicitly.
It would be interesting to analyze how such a solution affects robustness.

\section*{Acknowledgements}

SN is supported by the German Ministry for Education and Research as BIFOLD - Berlin Institute for the Foundations of Learning and Data (ref. 01IS18025A and ref. 01IS18037A).
RGK was supported by a grant from SAP Corporation.

\bibliography{sources}
\bibliographystyle{abbrvnat}

\clearpage
\appendix
\section{Theorem 1 does not apply to the MVAE}
\label{sec:app_mvae}

The MVAE \citep{wu2018multimodal} employs a product posterior inspired by the true posterior:
\begin{align}
    \q(\g|\{\bfx_m\}_{m=1}^{M}) \propto \p(\g) \prod_{m=1}^M \q(\g|\x_m).
\end{align}

In our experiments from \cref{sec:experiments}, we follow \citet{wu2018multimodal} and maximize the following three ELBOs:
\begin{align}
    L(\bftheta, \bfphi; \bfX) \coloneqq ELBO(\x_{1}, \x_2) + ELBO(\x_1) + ELBO(\x_2)
\end{align}
The ELBO for $M$ modalities is defined as:
\begin{align}
\begin{split}
    ELBO(\{\bfx_m\}_{m=1}^{M}) &\coloneqq
    \E_{\q(\g|\{\bfx_m\}_{m=1}^{M})} \left[ \log \frac{\p(\g)}{\q(\g|\{\bfx_m\}_{m=1}^{M})} \right]
    + 
    \sum_{m=1}^M \E_{\q(\g|\{\bfx_m\}_{m=1}^{M})} \left[
    \log \p(\x_m|\g)
    \right]\\
    &\leq     \log \p(\{\bfx_m\}_{m=1}^{M}),\\
\end{split}
\end{align}
Therefore, $\p(\x_m|\g)$ is always conditioned on $\x_m$ via the importance distribution, i.e., the model learns $p(\x_m|\{\bfx_i\}_{i=1}^{M})$ or $p(\x_m|\x_m)$.
This implies that the MVAE does not explicitly optimize $p(\x_{m\neq M}|\x_{M})$ for any $m\neq M$, i.e., \cref{theorem:main} does not apply to the MVAE.

\section{Additional experimental results}
\label{sec:app_exps}

\begin{wrapfigure}{r}{0.4\textwidth}
\vspace{-1cm}
  \begin{center}
    \includegraphics[width=\linewidth]{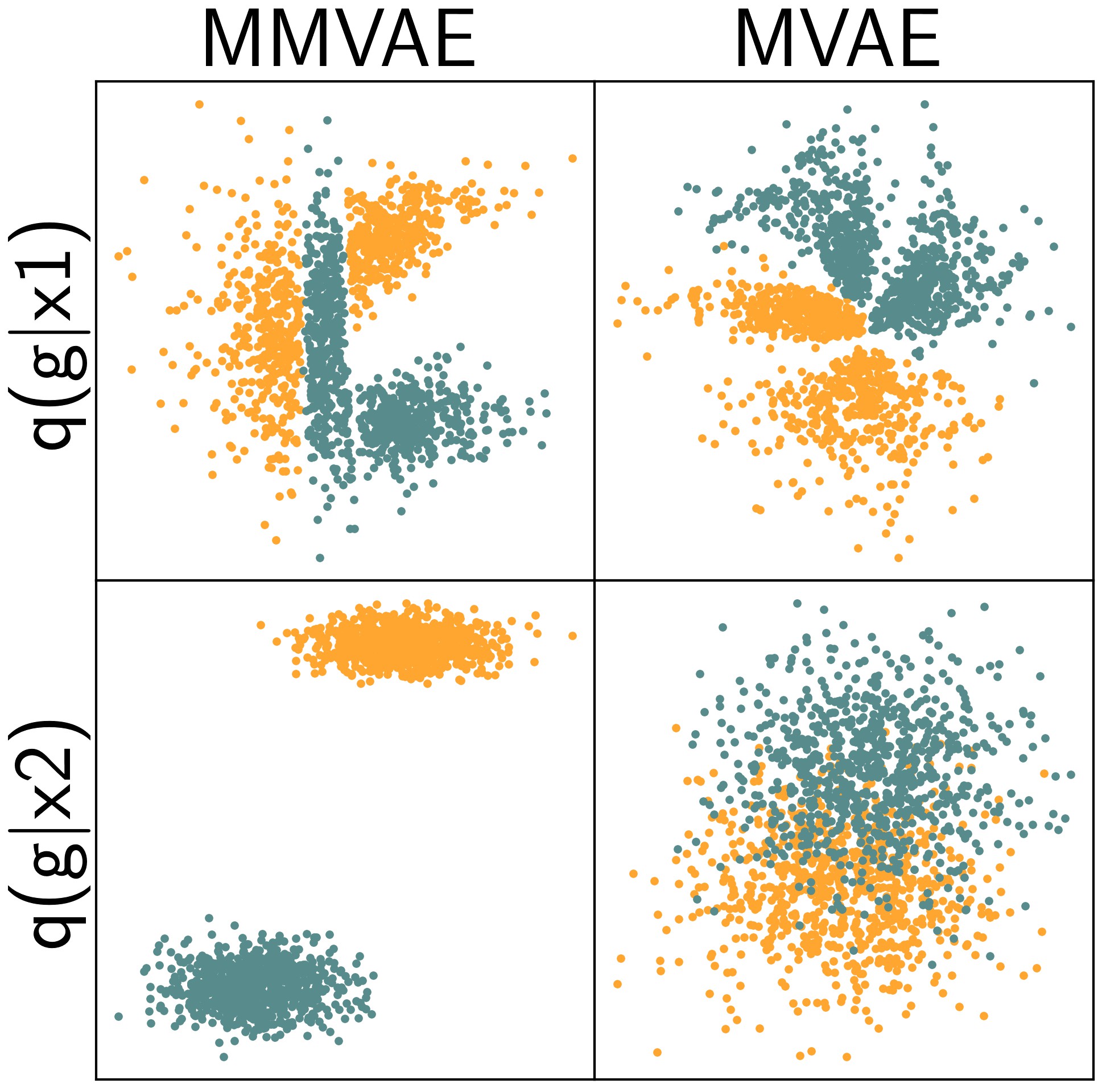}
  \end{center}
  \vspace{-0.2cm}
  \caption{\textbf{Marginal posteriors over the latent variable $\g$.}}
  \label{fig:latent_space}
\end{wrapfigure}

The solution $q(\g|\x_1)=q(\g|\x_2)$ can be helpful because it implies that samples from either posterior produce the same generative distribution for any modality.
\Cref{fig:latent_space} indicates that the MVAE aligns these marginal posteriors better than the MMVAE, which possibly explains the MVAE's better generative capability in \cref{fig:model_samples}.
\Cref{fig:model_samples} further exposes that even the MVAE struggles to represent the data perfectly.
Its latent representations from \cref{fig:latent_space} reveal that the model produces some overlap between the class manifolds of the marginal posteriors for the second modality -- possibly in an attempt to fit the isotropic Gaussian prior $p(\g)$.
We assume that this struggle is caused by the fact that there are just two unique label datapoints.

\end{document}